%% file: IJCNN2021.tex
\documentclass[conference]{IEEEtran}
\IEEEoverridecommandlockouts
\usepackage{cite}
\usepackage{amsmath,amssymb,amsfonts}
\usepackage{algorithmic}
\usepackage{graphicx}
\usepackage{xcolor}
\def\BibTeX{{\rm B\kern-.05em{\sc i\kern-.025em b}\kern-.08em
    T\kern-.1667em\lower.7ex\hbox{E}\kern-.125emX}}
    
\usepackage{xcolor}
\usepackage{graphicx}
\usepackage{multirow}
\usepackage{url}
\usepackage{xcolor}
\usepackage{graphicx}
\usepackage{arydshln}
\usepackage{microtype}
\usepackage{subfigure}
\usepackage[switch]{lineno}

\newcommand{\m}[1]{\mathbf{#1}}
\newcommand{\ms}[1]{\boldsymbol{#1}}
\newcommand{\tabincell}[2]{\begin{tabular}{@{}#1@{}}#2\end{tabular}}
\newcommand{\tabitem}{~~\llap{\textbullet}~~}

\usepackage{amsmath}
\usepackage{amsthm}
\usepackage{amssymb}
\usepackage{booktabs}
\usepackage{algorithm}
\usepackage{algorithmic}
\usepackage{subfigure}

\usepackage{xcolor}
\usepackage{graphicx}
\usepackage{multirow}
\usepackage{xcolor}
\usepackage{graphicx}
\usepackage{arydshln}
\usepackage{microtype}
\usepackage[switch]{lineno}

\begin{document}

\title{Discrete Auto-regressive Variational Attention Models for Text Modeling\\
}

\author{\IEEEauthorblockN{Xianghong Fang\IEEEauthorrefmark{1}\IEEEauthorrefmark{2},
Haoli Bai\IEEEauthorrefmark{1}\IEEEauthorrefmark{2}, Jian  Li\IEEEauthorrefmark{2}, 
Zenglin Xu\IEEEauthorrefmark{3}, 
Michael Lyu\IEEEauthorrefmark{2}, 
Irwin King\IEEEauthorrefmark{2}}
\IEEEauthorblockA{\IEEEauthorrefmark{2}Department of Computer Science and Engineering,
The Chinese University of Hong Kong\\
\IEEEauthorblockA{\IEEEauthorrefmark{3}School of Computer Science and Engineering, Harbin Institute of Technology, Shenzhen}
Email: xianghong\_fang@163.com, \{hlbai, jianli, lyu, king\}@cse.cuhk.edu.hk,
xuzenglin@hit.edu.cn\\
\IEEEauthorrefmark{1}: Equal Contribution\\
}
}

\maketitle

\begin{abstract}
Variational autoencoders~(VAEs) have been widely applied for text modeling. In practice, however, they are troubled by two challenges: information underrepresentation and posterior collapse. The former arises as only the last hidden state of LSTM encoder is transformed into the latent space, which is generally insufficient to summarize the data. The latter is a long-standing problem during the training of VAEs as the optimization is trapped to a disastrous local optimum. In this paper, we propose Discrete Auto-regressive Variational Attention Model (DAVAM) to address the challenges. Specifically, we introduce an auto-regressive variational attention approach to enrich the latent space by effectively capturing the semantic dependency from the input. We further design discrete latent space for the variational attention and mathematically show that our model is free from posterior collapse. Extensive experiments on language modeling tasks demonstrate the superiority of DAVAM against several VAE counterparts. Code will be released.
\end{abstract}


\input{sections/introduction.tex}

\input{sections/background.tex}

\input{sections/methods.tex}

\input{sections/experiment.tex}
\input{sections/related.tex}
\input{sections/conclusion.tex}
\input{sections/ack.tex}

\bibliographystyle{./IEEEtran}
\bibliography{IJCNN2021}

\end{document}

%% file: sections/introduction.tex
\section{Introduction}


As one of the representative deep generative models, variational autoencoders~(VAEs)~\cite{Kingma2013AutoEncodingVB} have been widely applied in text modeling~\cite{chung2015recurrent,zhang2016variational,su2018variational,Wang2019NeuralGC,Li2019ASE,bai2018neural,liu2018structured,liu2017stochastic}. Given input text $\m x \in \mathcal{X}$, VAEs learn the variational posterior $q_{\phi}(\m z|\m x)$ through the encoder and reconstruct output $\hat{\m x}$ from latent variables $\m z$ via the decoder $p_{\theta}(\m x|\m z)$. 
Both encoder and decoder are usually implemented by deep recurrent networks such as LSTMs~\cite{hochreiter1997long} in text modeling. Despite the success of VAEs, two long-standing challenges exist for such variational models: information underrepresentation and posterior collapse.

\begin{figure}[t]
    \centering
    \includegraphics[width=0.45\textwidth]{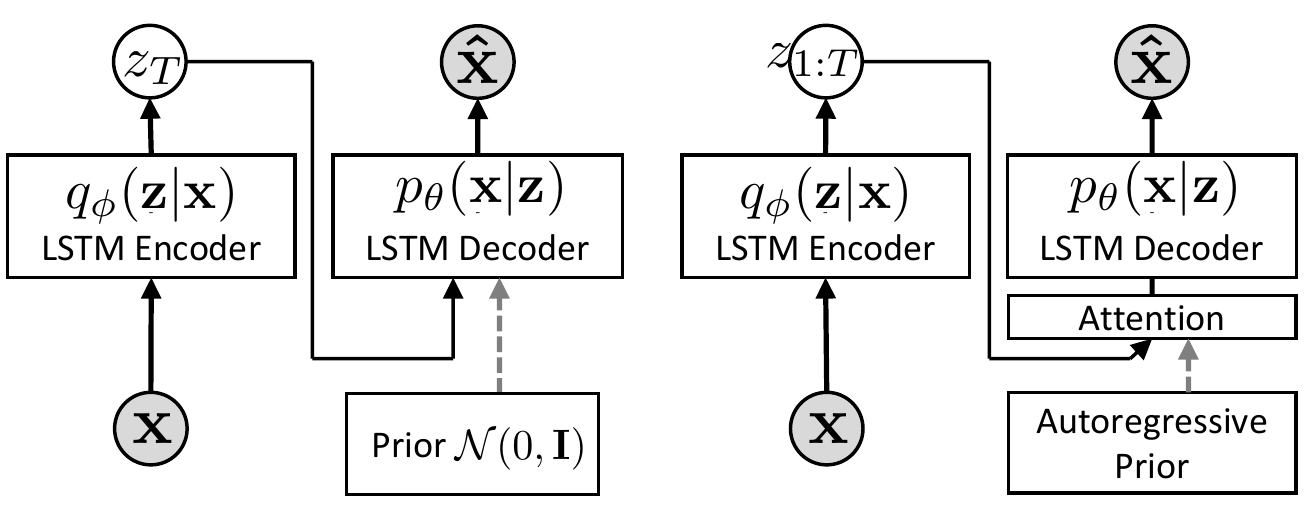}
    \caption{Illustration of conventional VAEs (left) and our proposed auto-regressive variational attention models (right).}
    \label{fig:intro}
\end{figure}

The challenge of information underrepresentation refers to the limited expressiveness of the latent space $\m z$. As shown in the left of Figure~\ref{fig:intro}, current VAEs build a single latent variable $\m z = z_T$ based on the last hidden state of LSTM encoder~\cite{Fu2019CyclicalAS,He2019LaggingIN,Wang2019NeuralGC,Li2019ASE}. However, this is generally insufficient to summarize the input sentence~\cite{bahuleyan2017variational}. Thus the generated sentences from the decoder are often poorly correlated. Notably, the sequence of encoder hidden states reflects the semantic dependency of the input sentence, and the whole hidden context may benefit the generation. Therefore, a potential solution is to enhance the representation power of VAEs via the attention mechanism~\cite{Bahdanau2014NeuralMT,Luong2015EffectiveAT}, a superior component in discriminative models. However, the attention module cannot be directly deployed in generative models like VAEs, as the attentional context vectors are hard to compute from randomly sampled latent variables during the generation phase.

Posterior collapse is another well-known problem during the training of VAEs~\cite{Bowman2015GeneratingSF}. It occurs as the variational posterior $q_{\phi}(\m z|\m x)$ converges to the prior distribution $p(\m z)$, thus the decoder receives no supervision from the input $\m x$.
Previous efforts alleviate this issue by either annealing the KL divergence term~\cite{Bowman2015GeneratingSF,Kingma2017ImprovedVI,Fu2019CyclicalAS}, revising the model~\cite{Yang2017ImprovedVA,Semeniuta2017AHC,Xu2018SphericalLS}, or modifying the training procedure~\cite{He2019LaggingIN,Li2019ASE}. 
Nevertheless, they primarily focus on a single latent variable for language modeling, which still suffer from the information underrepresentation as mentioned before. To derive more powerful latent space, the challenge of posterior collapse should be carefully handled.

In this paper, we propose Discrete Auto-regressive Variational Attention Model~(DAVAM) to address the aforementioned challenges.
First, to mitigate the information underrepresentation of VAEs, we introduce a variational attention mechanism together with an auto-regressive prior (dubbed as \emph{auto-regressive variational attention}). The variational attention assigns a latent sequence $\m z=z_{1:T}$ over each encoder hidden state to capture the semantic dependency from the input, as is shown in the right of Figure~\ref{fig:intro}. During the generation phase, the auto-regressive prior generates well-correlated latent sequence for computing the attentional context vectors.
Second, we utilize \emph{discrete latent space} to tackle the posterior collapse in VAEs. We show that the proposed auto-regressive variational attention models, when armed with conventional Gaussian distribution, face high risks of posterior collapse.
Inspired by the recently proposed Vector Quantized Variational Autoencoder~(VQVAE)~\cite{Oord2017NeuralDR,Razavi2019GeneratingDH}, we design a discrete latent distribution over the variational attention mechanism.
By analyzing the intrinsic merits of discreteness, we demonstrate that our design is free from posterior collapse regardless of latent sequences length. Consequently, the representation power of DAVAM can be significantly enhanced without posterior collapse.

We evaluate DAVAM on several benchmark datasets on language modeling. The experimental results demonstrate the superiority of our proposed method in text generation over its counterparts.

Our contributions can thus be summarized as:
\begin{enumerate}
	\item To the best of our knowledge, this is the first work that proposes \emph{auto-regressive variational attention} to improve VAEs for text modeling, which significantly enriches the information representation of latent space.
	\item We further design \emph{discrete latent space} for the proposed variational attention, which effectively addresses the posterior collapse issue during the optimization.
\end{enumerate}


%% file: sections/background.tex
\section{Background}
\subsection{Variational Antoencoders for Text Modeling}
Variational Autoencoders~(VAEs)~\cite{Kingma2013AutoEncodingVB} are a well known class of generative models. Given sentences $\m x= x_{1:T}$ with length $T$, we seek to infer latent variables $\m z$ that explain the observation. To achieve this, we need to maximize the marginal log-likelihood $\log p_\theta(\m x)$, which is usually intractable due to the complex posterior $p(\m z|\m x)$. Consequently an approximate posterior $q_{\phi}(\m z|\m x)$ (i.e. the \textit{encoder}) is introduced, and the evidence lower bound (ELBO) of the marginal likelihood is maximized as follows:
\begin{align}
\log p_\theta (\m x) \geq  & \underbrace{\mathbb{E}_{\m z \sim q_{\phi}(\m z|\m x)} [\log  p_{\theta}(\m x|\m z)]}_\textrm{reconstruction loss} \nonumber \\
& \hspace{4ex} - \underbrace{ D_{KL}(q_{\phi}(\m z|\m x) \Vert p(\m z))}_\textrm{KL divergence},
\label{eq:elbo}
\end{align}
where $p_{\theta}(\m x|\m z)$ represents likelihood function conditioned on $\m z$, also known as the \textit{decoder}. In the context of text modeling, both encoder and decoder are usually implemented by deep recurrent models such as LSTMs~\cite{hochreiter1997long}, parameterized by $\phi$ and $\theta$ respectively.



\subsection{Challenges}

\paragraph{Information Underrepresentation} 
Information underrepresentation is a common issue in applying VAEs for text modeling. Conventional VAEs build latent variables based on the last hidden state of LSTM encoder, i.e. $\m z=z_T$. During the decoding process, we first sample $z_T$, from which new sentences $\hat{\m x}=\hat{x}_{1:\hat{T}}$ can be generated:
\begin{equation}
p(\hat{\m x}|\m z) = p_{\theta}(\hat{x}_1|z_T) \prod_{t=2}^{\hat{T}} p_{\theta}(\hat{x}_{t}|\hat{x}_{t-1}, z_T),
\label{eq:decoder_generation}
\end{equation}
where $\hat{T}$ is the length of reconstructed sentence $\hat{\m x}$.
However, the representation of $z_T$ is generally insufficient to summarize the semantic dependencies in $\m x$, and thus deteriorates the reconstruction.

\paragraph{Posterior Collapse}
Posterior collapse usually arises as $D_{KL}(q_{\phi}(\m z|\m x) \Vert p(\m z))$ diminishes to zero, where the local optimal gives $q_{\phi}(\m z|\m x) = p(\m z)$. 
Posterior collapse happens inevitably as the ELBO contains both the reconstruction loss $\mathbb{E}_{\m z \sim q_{\phi}(\m z|\m x)}[\log  p_{\theta}(\m x|\m z)]$ and the KL-divergence $D_{KL}(q_{\phi}(\m z|\m x) \Vert p(\m z))$, as shown in Equation~(\ref{eq:elbo}).
When posterior collapse happens, $\m x$ becomes independent of $\m z$ as $p(\m x)p(\m z) = p(\m x)q_{\phi}(\m z| \m x) = p(\m x)\frac{p(\m x, \m z)}{p(\m x)} = p(\m x, \m z)$. Therefore, the encoder learns a data-agnostic posterior without any information from $\m x$, while the decoder fails to perform valid generation but purely based on random noise.






%% file: sections/methods.tex
\section{Methodology}
\begin{figure*}
	\centering
	\includegraphics[width=0.95\textwidth]{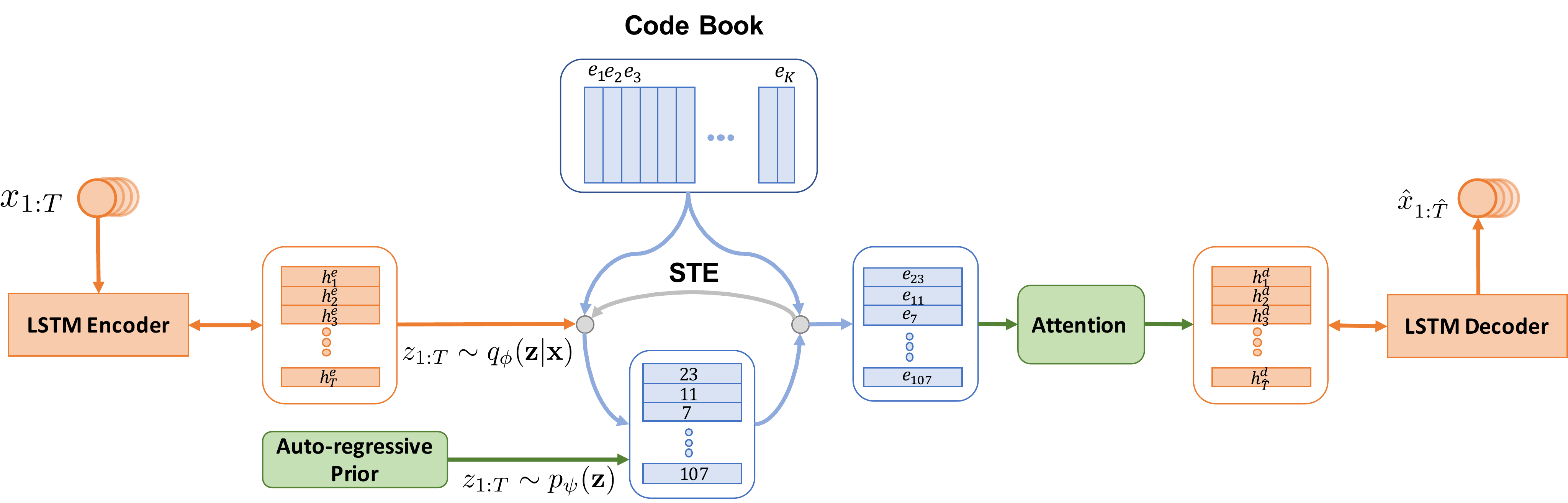}
	\caption{The overall architecture of the proposed DAVAM. Given observations  $\m x=x_{1:T}$, the encoder hidden states $h^e_{1:T}$ are quantized to code book $\{e_k\}_{k=1}^K$ based on index sequence $z_{1:T}$ from the posterior. The quantized hidden states $e_{z_{1:T}}$ are then forwarded to the attention module with decoder hidden states $h_{1:\hat{T}}^d$.
	During back-propagation, the gradients of $e_{z_{1:T}}$ are directly copied to $h^e_{1:T}$ with STE.
	To generate new sentences from DAVAM, we start from the auto-regressive prior to sample a new latent sequence $z_{1:T}$. The sequence $z_{1:T}$ is then used to index the code book for attention computation during decoding.}
	\label{fig:vam model}
\end{figure*}

We now present our solutions to address the aforementioned challenges. In order to enrich the latent space, we propose an auto-regressive variational attention model to capture the semantic dependencies in the input space. We first instantiate variational attention with the Gaussian distribution and show that it suffers from posterior collapse. Then to solve the challenge, we further discretize the latent space with one-hot categorical distribution, leading to discrete auto-regressive variational attention models~(DAVAM), as illustrated in Figure~\ref{fig:vam model}. We carefully analyze the superiority of DAVAM to avoid posterior collapse.


\subsection{Gaussian Auto-regressive Variational Attention Models}
\label{sec:vam}
To enrich the representation of latent space $\m z$, we seek to incorporate the attention mechanism into VAEs. Specifically, we denote the encoder hidden states as $ h^e_{1:T}$, and the decoder hidden states as $h^d_{1:\hat{T}}$. We build a latent sequence $\m z = z_{1:T}$ upon encoder hidden states $ h^e_{1:T}$. 
To facilitate such variational attention model, one can choose the conventional Gaussian distribution~\cite{Kingma2013AutoEncodingVB} for variational posteriors, i.e. $q(\m z|\m x)=\prod_{t=1}^{T}q(z_t| \m x)$ where $q_\phi(z_t|\m x) = \mathcal{N}(\mu_t, \sigma_t I)$. We name the resulting model as Gaussian Auto-regressive Variational Attention Model~(GAVAM). 


Given $z_{1:T}$, similar to attention-based sequence-to-sequence~(seq2seq) models~\cite{Bahdanau2014NeuralMT}, the attentional context vectors $c_i$ and scores at $i$-th decoding step can be computed by 
\begin{align}
\label{eq:gvam_attention}
c_i = \sum_{t=1}^{T} \alpha_{i, t} z_{t}, \hspace{1ex} \alpha_{i, t} = \frac{\exp(\tilde{\alpha}_{i, j})}{\sum_{j=1}^{T}\exp(\tilde{\alpha}_{i, j})},
\end{align}
where $\tilde{\alpha}_{i, t}=v^{\top}\tanh(W_e z_t + W_d h^d_{i-1} + b)$ is the unnormalized score, $t\in \{1,2,...,T\}$ is the encoder time step, and $W_e, W_d$ are the corresponding parameters. By taking $c_i$ as extra input to the decoder, the generation process is reformulated as
\begin{align}
p(\hat{\m x}| \m c, \m z) = p(\hat{x}_1| c_1, z_{1:T})\prod_{\hat{t}=2}^{\hat{T}} p(\hat{x}_{\hat{t}}|\hat{x}_{\hat{t}-1}, c_{\hat{t}}, z_{1:T}).
\nonumber
\end{align}
Unlike Equation~\ref{eq:decoder_generation}, at each time step, the decoder receives supervision from the context vector, which is a weighted sum of the latent sequence $z_{1:T}$. Consequently, the variational posterior  $q_\phi(z_{1:T}|\m x)$ encodes the semantic dependency from the observations, such that the issue of information underrepresentation can be effectively mitigated.





\paragraph{Auto-regressive Prior} A key difference between variational auto-regressive attention models and conventional VAEs is the choice of a prior distribution. During the generation, the latent sequence $z_{1:T}$ are sampled from the prior unconditionally, and are then fed to the 
attention module together with $h_{1:\hat{T}}^d$. The most adopted prior $\mathcal{N}(0, I)$, however, is non-informative to generate
\emph{well-correlated} latent sequence for the attention as that during training. Therefore the decoder receives no informative supervision that gives reasonable generation.


To solve that, we deploy an auto-regressive prior $p_\psi(z_{1:T}) = p_\psi(z_1)\prod_{t=2}^T p_\psi(z_t|z_{1:t-1})$ parameterized by $\psi$ to capture the underlying semantic dependencies. Specifically, we take $p_{\psi}(z_t|z_{1:t-1}) = \mathcal{N}(\hat{\mu}_t, \hat{\sigma}_t I)$, where $(\hat{\mu}_t, \hat{\sigma}_t I)$ is produced by a PixelCNN, a superior model in learning sequential data~\cite{Oord2016ConditionalIG}.


\paragraph{Posterior Collapse in GAVAM}
The training of GAVAM can be easily troubled by posterior collapse due to two aspects. To see this, similar to Equation~\ref{eq:elbo}, the minimization of the ELBO now can be written as:
\begin{align}
    \label{eq:GAVAM_elbo}
    \min_{\phi,\theta,\psi} & - \mathbb{E}_{ z_{1:T}\sim q_{\phi}} 
    [\log p_{\theta}( \m x| z_{1:T})] \\
    & \hspace{2ex}+ \sum_{t=1}^{T}D_{KL}(q_{\phi}({z_t}|{\m x}) \Vert p_\psi({z_t|z_{1:t-1}})). \nonumber
\end{align}
On the one hand, the KL divergence scales linearly to the sequence length of $T$, which makes the training unstable across different input lengths. 
On the other hand, and more seriously, both $\phi$ and $\psi$ are used to minimize the KL divergence, which can easily trap the learned posteriors. To demonstrate this, for example, the KL divergence between two Gaussian distributions can be written as:
\begin{align}
\label{eq:gaussian_kl}
& \sum_{t=1}^{T}D_{KL}(q_{\phi}({z_t}|{\m x}) \Vert p_\psi({z_t|z_{1:t-1}})) \\
& = \sum_{t=1}^T \sum_{d=1}^D \frac{1}{2} (\log \frac{\hat{\sigma}_{td}^2}{\sigma_{td}^2} - 1 + \frac{\sigma_{td}^2 + (\hat{\mu}_{td} - \mu_{td})^2}{\hat{\sigma}_{td}^2}), \nonumber
\end{align}
where $D$ is the latent dimension of $z_t$. Whenever $\sigma^2_{td} \rightarrow \hat{\sigma}^2_{td}$ and $\mu_{td} \rightarrow \hat{\mu}_{td}$ before $q_\phi(z_{1:T}|\m x)$ encodes anything from $\m x$, both $q_\phi(z_{1:T}|\m x)$ and $p_\psi(z_{1:T})$ get stuck in local optimal and learn no semantic dependency for reconstruction.
\subsection{Discrete Auto-regressive Variational Attention Models}
Inspired by recent studies~\cite{Oord2017NeuralDR,Roy2018TheoryAE} that demonstrate the promising effects of \emph{discrete latent space}, we explore its potential in handling posterior collapse over the variational attention, leading to discrete auto-regressive variational attention model~(DAVAM).



Specifically, we introduce a code book $\{e_k\}_{k=1}^K$ with size of $K$, where each $e_k$ is a vector in the latent space. We expect the combination of code book can represent the semantic dependency from observed sentence $\m x$. We now substitute the Gaussian distributed $z_{1:T}$ with discrete indices over code book that follows one-hot categorical distribution:
\begin{equation}
\label{eq:vq_distribution}
q_\phi(z_t=k|\m x) =
\begin{cases}
1 & k=\arg\min_{j}\|h^e_t-e_j\|_2\\
0 & \text{otherwise}
\end{cases}.
\end{equation}
Given index $z_t$, we transform the the encoder hidden state $h_t^e$ to the nearest $e_{z_t}$. Then we use $e_{z_t}$ instead of $z_t$ in Equation~\ref{eq:gvam_attention} to compute attention scores $\alpha_t$ and the context vectors $c_i$.

Correspondingly, as $z_{1:T}$ are discrete indices, we assign categorical distribution for the auto-regressive prior, i.e., $p_{\psi}(z_t|z_{1:t-1}) = \mathrm{Cat}(\gamma_t)$. The categorical parameter can be obtained from the PixelCNN model given historical records $z_{1:t-1}$, i.e, $\gamma_t=\mathrm{PixelCNN}_{\psi}(z_{1:t-1}) \in [0,1]^K$.



\paragraph{Advantages of Discreteness}
Thanks to the nice properties of discreteness, the optimization of DAVAM does not suffer from posterior collapse. 
Specifically, the KL divergence of DAVAM can be written as:
\begin{align}
    \label{eq:vqvae_kl}
    & \sum_{t=1}^T D_{KL}(q_\phi( z_t | \m x)||p_\psi({z_t|z_{1:t-1}})) \\
    & = - \sum_{t=1}^T \Big[ H(q_\phi( z_t)) + \sum_{k=1}^K 1_{(z_t=k)} \log \gamma_{t,k} \Big] \nonumber \\
    &= - \sum_{t=1}^T \Big[ 0 + \log \gamma_{t,z_t}\Big], \nonumber
\end{align}
where the third line is obtained with the entropy $H(q_\phi( z_t)) = -1 \log 1 - 0 \log 0 \rightarrow 0$. It can be found that  $D_{KL}(q_{\phi}(z_{1:T}|\m x) \Vert p_{\psi}(z_{1:T}))$ is no longer relevant to posterior parameters $\phi$. 
Consequently, the update of variational posterior $q_\phi(z_{1:T}|\m x)$ does not rely on the prior but is determined purely by the reconstruction term. Therefore minimization of KL divergence will not lead to posterior collapse.




\subsection{Model Training}
We first train the variational posterior $q_{\phi}(\m z|\m x)$ to convergence when the latent sequence $z_{1:T}$ effectively captures the semantic dependency from input $\m x$. Then we train the auto-regressive prior $p_{\psi}(\m z)$ to mimic the learned posterior, to facilitate well-correlated sequence during generation. Therefore, the training of the proposed DAVAM involves two stages, described in detail as follows:



\paragraph{Stage one}
We follow the standard paradigm to minimize the ELBO of DAVAM. As shown in Equation~\ref{eq:vqvae_kl}, since KL divergence is neither relevant to $\phi$ nor $\theta$, only the reconstruction term should be concerned.
In the meanwhile, as the latent variables $z_{1:T}$ are determined based on Euclidean distances between $h^e_{1:T}$ and code book $\{e_k\}_{k=1}^K$, we further regularize them to stay close via a Frobenius norm. The training objective for stage one is
\begin{equation}
    \label{eq:vqvae_loss}
    \min_{\theta, \phi} -\mathbb{E}_{q_{\phi}} 
    [\log p_{\theta}( \m x| z_{1:T})] + \beta\sum_{t=1}^{T}\| h_t^e - \mathrm{{sg}}(e)\|_F^2,
\end{equation}
where $\beta$ is the regularizer, and $\mathrm{{sg(\cdot)}}$ stands for stop-gradient operation.
Note that the quantization in Equation~\ref{eq:vq_distribution} is non-differentiable. To allow the back-propagation algorithm to proceed, we adopt the widely employed straight through estimator~(STE)~\cite{bengio2013estimating} to copy gradients from $e_{z_t}$ to $h_{t}^{e}$, as is shown in Figure~\ref{fig:vam model}. 

For the update of code book $\{e_k\}_{k=1}^K$, 
we first apply K-means algorithm to calculate the average over all latent variables $h_{1:T}^e$ that are closest to $\{e_k\}_{k=1}^K$, and then take exponential moving average over the mini-batch update of the code book so as to stabilize the optimization.

\paragraph{Stage Two}
After the convergence of DAVAM, we resort to update the auto-regressive prior $p_\psi( z_t|z_{1:t-1})$. To mimic the semantic dependency in the learned posterior $q_\phi (z_{1:T}|\m x)$, the prior is supposed to fit the latent sequence $z_{1:T}\sim q_{\phi}(z_t|\m x)$. This can be realized by the minimizing their KL-divergence w.r.t. $\psi$ as
\begin{equation}
    \min_{\psi} \sum_{t} D_{KL}(q_\phi( z_t| \m x)||p_\psi( z_t|z_{1:t-1})),
\end{equation}
which can be simplified to the cross-entropy loss between $z_{1:T}$ and $\gamma_{1:T}$ according to Equation~(\ref{eq:vqvae_kl}).

%% file: sections/experiment.tex
\section{Experiments}
We verify advantages of the proposed DAVAM on language modeling tasks, and testify how well can it generate sentences from random noise. Finally, we conduct a set of further analysis to shed more light on DAVAM. Code implemented in PyTorch is available at \url{https://github.com/sunset-clouds/DAVAM}.


\subsection{Experimental Setup}

We take three benchmark datasets of language modeling for verification: Yahoo Answers ~\cite{Xu2018SphericalLS}, Penn Tree~\cite{Marcus1993BuildingAL}, and a down-sampled version of SNLI~\cite{bowman2015large}. A summary of dataset statistics is shown in Table~\ref{table:datasets}.

\begin{table}[h]
\caption{Dataset statistics.}
\resizebox{0.48\textwidth}{!}{
\begin{tabular}{c|cccc}
\hline
Datasets &Train Size&Val Size &Test Size & Avg Len\\\hline
Yahoo & 100,000 & 10,000 & 10,000& 78.7\\
PTB & 42,068 & 3,370 & 3,761  & 23.1 \\
SNLI & 100,000 & 10,000 & 10,000  & 9.7\\
\hline
\end{tabular}}
\label{table:datasets}
\end{table}


\begin{table*}[t]
\begin{center}
\caption{Results of language modeling on Yahoo, PTB, and SNLI Datasets. For both Rec and PPL, the lower the better. For KL, a small value indicates the posterior collapse, but this is not a issue for DAVAM (marked by ``\underline{\ \ \ }").}
\resizebox{0.9\textwidth}{!}{
\begin{tabular}{l||l|c|c|c|c|c|c|c|c|c}
\multirow{2}{*}{\#} & \multirow{2}{*}{Methods} & \multicolumn{3}{|c}{Yahoo} 
& \multicolumn{3}{|c}{PTB} & \multicolumn{3}{|c}{SNLI}\\
\cline{3-11}
& & Rec$\downarrow$ & PPL$\downarrow$ & KL & Rec$\downarrow$ & PPL$\downarrow$ & KL & Rec$\downarrow$ & PPL$\downarrow$ & KL\\\hline
1 & LSTM-LM & - & 60.75  & -  & - & 100.47  & - & - & 21.44  & - \\
2 & VAE & 329.10 & 61.52 & 0.00 &  101.27 & 101.39 & 0.00 & 33.08 & 21.67 & 0.04\\
3 & +anneal & 328.80 & 61.21 & 0.00  & 101.28  & 101.40 & 0.00 & 31.66 & 21.50 & 1.42\\
4 & +cyclic & 333.80 & 66.93 & 2.83 & 101.85 & 108.97 & 1.37 & 30.69 & 23.67 & 3.63\\
5 & +aggressive & 322.70 & 59.77 & 5.70 & 100.26 & 99.83 & 0.93 & 31.53 & 21.16 & 1.42\\
6 & +FBP  & 322.91 & 62.59 & 9.08 & 98.52 & 99.62 & 2.95 & 25.26 & 22.05 & 8.99\\
7 & +pretraining+FBP & 315.09  & 59.60 & 15.49 & 96.91 & 96.17 & 4.99 & 22.30 & 22.33 & 13.40 \\\hline
8 & GAVAM & 350.14 & 79.28 & 0.00 & 102.20 & 105.94 & 0.00 & 30.90 & 17.68 & 0.38    \\
9 & DAVAM-q (K=512) & 323.10 & 57.14 & \underline{0.33} & 95.83 & 79.24 & \underline{0.27} & 28.16 & 13.71 & \underline{0.12} \\
10 & DAVAM (K=128) & 303.65 & 45.07 & \underline{1.88} & 83.57 & 50.15 & \underline{2.23} & 16.11 & 5.58 & \underline{2.38}  \\
11 & DAVAM (K=512) & \textbf{259.68} & \textbf{26.61} & \underline{2.60} & \textbf{60.16} & \textbf{17.94} & \underline{3.12} & \textbf{10.85} & \textbf{3.52} & \underline{2.69}  \\
\end{tabular}}
\label{table:density_results}
\end{center}
\vspace{-2ex}
\end{table*}



\paragraph{Baselines}
We compare the proposed DAVAM against a number of baselines, including the classical LSTM-based Language Modeling-(LSTM-LM), vanilla VAE~\cite{Kingma2013AutoEncodingVB}, and its advanced variants: annealing VAE~\cite{Bowman2015GeneratingSF}, cyclic annealing VAE\footnote{\url{https://github.com/haofuml/cyclical_annealing}}~\cite{Fu2019CyclicalAS}, lagging  
VAE\footnote{\url{https://github.com/jxhe/vae-lagging-encoder}}~\cite{He2019LaggingIN}, Free Bits (FB)~\cite{Kingma2017ImprovedVI} and pretraining+FBP VAE\footnote{\url{https://github.com/bohanli/vae-pretraining-encoder}}~\cite{Li2019ASE}. All these baselines do not use the attention module in their architectures.

For ablation studies,  we further compare to 1) GAVAM, which takes Gaussian distribution instead of the one-hot categorical distribution over $z_{1:T}$ to verify the advantages of discreteness; 2) We also remove the attention mechanism (denoted as DAVAM-q) to test the effect of discreteness on the last latent variable $z_T$. 
3) Finally, to check the choice of prior, we replace the auto-regressive prior with uninformative Gaussian priors, which leads to variational attention models~(VAE+Attn) first proposed by~\cite{bahuleyan2017variational,bahuleyan2017variational}. %





\paragraph{Evaluation Metrics} We evaluate language modeling using three metrics: 1) Reconstruction loss (Rec) $\mathbb{E}_{\ms z \sim q_{\phi}(\ms{z} | \ms x)}[\log  p_{\theta}(\ms {x}|\ms{z})]$ that measures the ability to recover data from latent space; 2) Perplexity (PPL) measuring the capacity of language modeling; Both lower Rec and PPL give better models in general; and 3) KL divergence (KL) indicating whether posterior collapse occurs. 


\paragraph{Implementation}
For baselines, we keep the same hyper-parameter settings to pretraining+FBP VAE~\cite{Li2019ASE}, e.g., the dimension of latent space, word embeddings and hidden states of LSTM.
Since our latent variables are discrete, we cannot use importance weighted samples to approximate the reconstruction loss in Lagging VAE and pretraining+FBP VAE. 

For DAVAM and its ablation counterparts, we keep the same set of hyper-parameters. By default, we set the code-book size $K$ as $512$.
We first warm up the training for 10 epochs, and then gradually increase $\beta$ in Equation~(\ref{eq:vqvae_loss}) from 0.1 to $\beta_{max} = 5.0$, in a similar spirit to annealing VAE~\cite{Bowman2015GeneratingSF}. 
For all experiments, we use the SGD optimizer with initial learning rate $1.0$, and decay it until five counts if the validation loss does not decrease for $2$ epochs. For the auto-regressive prior, we use a 16-layer PixelCNN with one-dimensional convolution followed by residual connections.


\subsection{Experimental Results}

\begin{table*}[t]
\begin{center}
\caption{Sampled short, medium and long sentences as well as their GPT-2 PPL scores for measuring fluency. }
\footnotesize
\resizebox{0.98\textwidth}{!}{
\begin{tabular}{l|l|c}
\hline
Methods & Samples & PPL$\downarrow$ \\\hline
pretraining &  \tabitem [s] i can say what can be the least in terms also in any form of power stream [/s] & 5.97 \\
\cdashline{2-3}
\cdashline{2-3}
 +FBP &\tabitem [s] how you define yourself ? birth control. you will find yourself a dead because of your periods. [/s] & 5.43\\
\cdashline{2-3}
VAE & \tabitem [s] are they allowed to join (francisco) in \_UNK. giants in the first place.? check out other answers. & 5.21\\
& do you miss the economy and not taking risks in the merchant form, what would you tell? go to & \\
&the yahoo home page and ask what restaurants follow this one. [/s] &  \\
\hline
VAE+Attn
& \tabitem [s] explain some make coming you think and represents middle line girl coming. [/s]  & 8.83 \\
\cdashline{2-3}
\cdashline{2-3}
& \tabitem [s] who just live this in you to get usual usual help the idea for out to use guess guess thats UNK  & 7.16 \\
\cdashline{2-3}
& \tabitem [s]  is masterbating masterbating anyone hi fact fact forgive forgive forgive virgin chlorine does & 5.01\\
&'re hydrogen download 're whats 're does solve 're whats solve 2y germany germany monde&\\ 
&pourquoi 'm does fun pourquoi 'm 're 'm 're solve 'm does solve pourquoi 'm 'm 'm 'm solve 'm 'm [/s] & \\
\hline
GAVAM 
& \tabitem [s] generally do problems problems, do you have problems [/s]  & 6.57 \\
\cdashline{2-3}
\cdashline{2-3}
& \tabitem [s] plz can you get a UNK envelope in e ? for me for my switched for eachother i would switched to i & 6.50 \\
& i think  thats . i need to assume . [/s] & \\
\cdashline{2-3}
\cdashline{2-3}
& \tabitem [s] what to do, yoga is there to place and pa? i can definately, but that, you will the best, but the the only &5.18\\
&amount right? the best range, it though, to do n't do to be to be out, and  [/s] & \\
\hline
DAVAM & \tabitem [s] what is the meaning of time management? [/s] & 4.52 \\
\cdashline{2-3}
\cdashline{2-3}
(K=512) & \tabitem [s] what should you be thankful for thanksgiving dinner and how to get some money with a thanksgiving &4.25\\
&dinner? [/s] &  \\
\cdashline{2-3}
& \tabitem [s] is anyone willing to donate plasma if you are allergic to cancer or anything else? probably you can. & 3.87\\
&i've never done any thing but it is only that dangerous to kill bacteria. i have heard that it doesn't have & \\
&any effect on your immune system. [/s]  & \\
\bottomrule
\hline
\end{tabular}}
\label{table:sentences}
\end{center}
\end{table*}

\subsubsection{Language Modeling}
\label{sec:exp_lm}
To compare the representation of latent space, we first perform language modeling over the testing corpus of benchmark datasets, as shown in Table~\ref{table:density_results}.
Generally, the better the representation, the lower the Rec and PPL on observations.
For DAVAM and GAVAM, we average the KL divergence along the latent sequence to make them comparable to baselines that only have one latent variable.

\paragraph{Main Results} (Rows 1-7,10-11)
Comparing to baselines without variational attention, we find that 
our DAVAM achieves significantly better results on all three datasets, especially with larger code book size $K$.
For example, comparing to pretraining+FBP in row 7 on Yahoo Answers dataset, DAVAM with $K=512$ significantly reduces the reconstruction loss by $55.41$, and PPL is decreased by more than a half. 
Therefore, our DAVAM is more expressive to summarize observations comparing to baselines without attention modules. The success verifies that DAVAM can significantly enrich the latent representation of language modeling.

In terms of posterior collapse, both vanilla VAE and some variants suffer from this issue severely as the KL of Gaussian distribution diminishes nearly to 0. However, the KL of DAVAM does not indicate posterior collapse, but only reflects how well the auto-regressive prior mimic the posterior.


\paragraph{Ablation Studies} (Rows 8-11)
GAVAM performs less competitively on language modeling comparing to DAVAM. Moreover, its KL divergences are near or equal to 0. This leads to the posterior collapse and explains why it has the sub-optimal performance with the one-hot categorical distribution substituted.
In terms of DAVAM-q, it has no attention module and only learns the variational posterior with the last $z_T$, which naturally yields less competitive results against attention-based models.
However, DAVAM-q still outperforms a number of variants of VAE, as the posterior is free from the collapse thanks to discreteness.

\subsection{Language Generation From Scratch}
In this section, we dive into the ability of language generation from scratch, i.e. generating sentences directly from random noises. 
We study the generated languages from two perspective: quality and diversity.
Then we apply such input-free generation approach to data augmentation, which can be applied to improve the language models trained over limited corpus.


\begin{table}[t]
    \centering
    \vspace{-0.15in}
    \caption{GPT-2 perplexity scores ($\downarrow$) with standard deviation ($\pm$) for generation quality (row 1-5), and the reconstruction loss
($\downarrow$) on Yahoo dataset for language modeling (last row).}
    \resizebox{0.42\textwidth}{!}{
    \begin{tabular}{c||c|c|c}
    Length & \tabincell{c}{VAE+Attn} & GAVAM & DAVAM 
    \\\hline
         10 & $7.75_{\pm 1.71}$ & $7.32_{\pm 1.64}$  & $\mathbf{6.58_{\pm 1.17}}$ \\
         20 & $7.04_{\pm 1.46}$ & $6.94_{\pm 1.40}$  & $\mathbf{6.49_{\pm 1.09}}$ \\
         30 & $6.70_{\pm 1.31}$ & $6.75_{\pm 1.38}$  &  $\mathbf{5.97_{\pm 0.87}}$\\
         40 & $6.54_{\pm 1.50}$ & $6.28_{\pm 1.36}$  & $\mathbf{5.79_{\pm 0.81}} $ \\
         50 & $6.42_{\pm 1.61}$ & $6.10_{\pm 1.33}$  & $\mathbf{5.55_{\pm 0.97}}$ \\
         \hline
         Rec & $\mathbf{10.85}$ & $350.14$ & $259.68$
    \end{tabular}}
    \label{tab:gpt2_scores}
\end{table}

\subsubsection{Generation Quality}
We first visualize the sentences generated by Pretraining+FBP VAE, VAE+Attn, GAVAM, and DAVAM, along with their fluency scores (PPL).
We adopt a pre-trained GPT-2~\cite{radford2019language}~\footnote{\textrm{https://huggingface.co/transformers/model\_doc/gpt2.html}} as the fluency evaluator, which takes the generated sentences as input and returns the corresponding perplexity scores~(PPL).

From Table~\ref{table:sentences}, we find that VAE+Attn can hardly generate well-correlated latent sequences from uninformative Gaussian distributions as a result of non-autoregressive prior.
On the other hand, GAVAM is armed with the auto-regressive prior and shows more readable sentences. Nevertheless, it suffers from poor semantic meanings due to posterior collapse, as previously shown in Table~\ref{table:density_results}.
Finally, DAVAM can produce sentences with interpretable meanings and better fluency scores, even when the sequence length is long. This suggests the discrete latent sequence combined with auto-regressive prior enjoys unique advantages in language generation from scratch.

\begin{figure}[t]
	\centering
	\vspace{-0.15in}
	\includegraphics[width=0.4\textwidth]{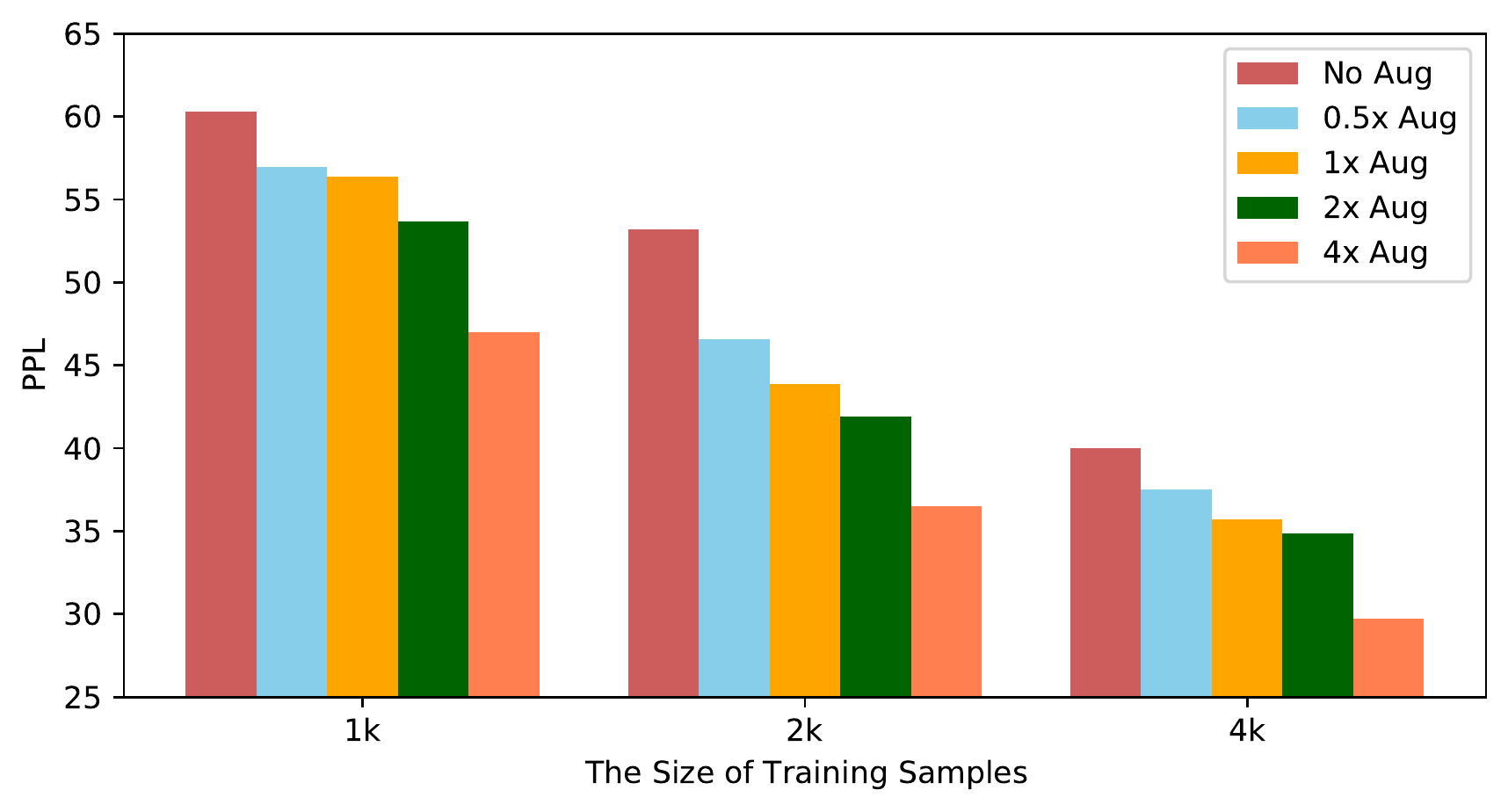}
	\hspace{-0.1in}
	\caption{The perplexity scores under different sizes of augmented training sentences on SNLI dataset.}
	\vspace{-2ex}
	\label{fig:barchart}
\end{figure}

\begin{table*}[h!]
	\begin{center}
	    \vspace{-0.1in}
		\caption{The generation diversity scores evaluated by entropy (Ent.), distinct unigrams (Dist-1) and bigrams (Dist-2).}
		\resizebox{1.0\textwidth}{!}{
			\begin{tabular}{c|c|c|c|c|c|c|c|c|c|c|c|c}
				\multirow{2}{*}{Length}  
				& \multicolumn{3}{|c}{pretraining+FBP VAE} 
				& \multicolumn{3}{|c}{VAE+Attn}
				& \multicolumn{3}{|c}{GAVAM} 
				& \multicolumn{3}{|c}{DAVAM}\\
				\cline{2-13}
				& Ent. $\uparrow$ & Dist-1$\uparrow$ & Dist-2$\uparrow$ & Ent. $\uparrow$ & Dist-1$\uparrow$ & Dist-2$\uparrow$ & Ent. $\uparrow$ & Dist-1$\uparrow$ & Dist-2$\uparrow$& Ent. $\uparrow$ & Dist-1$\uparrow$ & Dist-2$\uparrow$ \\\hline
				10 & 5.41 & 0.412 & 0.853 & 5.09 & 0.366 & 0.792 & 4.78 & 0.288 & 0.724 & 5.00 & 0.366 & 0.844 \\
				20 & 5.48 & 0.355 & 0.836  & 5.12 & 0.243 & 0.649 & 4.80 & 0.212 & 0.636 & 5.10 & 0.249 & 0.702 \\
				30 & 5.57 & 0.301 & 0.798 & 4.70 & 0.166 & 0.487 & 4.60 & 0.150 & 0.503 & 5.18 & 0.210 & 0.655 \\
				40 & 5.55 & 0.252 & 0.756 & 4.10 & 0.110 & 0.372 & 4.19 & 0.113 & 0.401 & 5.34 & 0.188 & 0.646 \\
				50 & 5.69 & 0.249 & 0.765 & 3.92 & 0.089 & 0.326 & 4.02 & 0.093 & 0.354 & 5.33 & 0.173 & 0.611 
		\end{tabular}}
		\label{table:generation_diversity}
	\end{center}
	\vspace{-3ex}
\end{table*}

\begin{figure*}[t]
	\centering
	\label{fig:ablation}
	\subfigure[Code book size $K$] { 
		\label{fig:two_stage}     
		\includegraphics[width=0.233\textwidth]{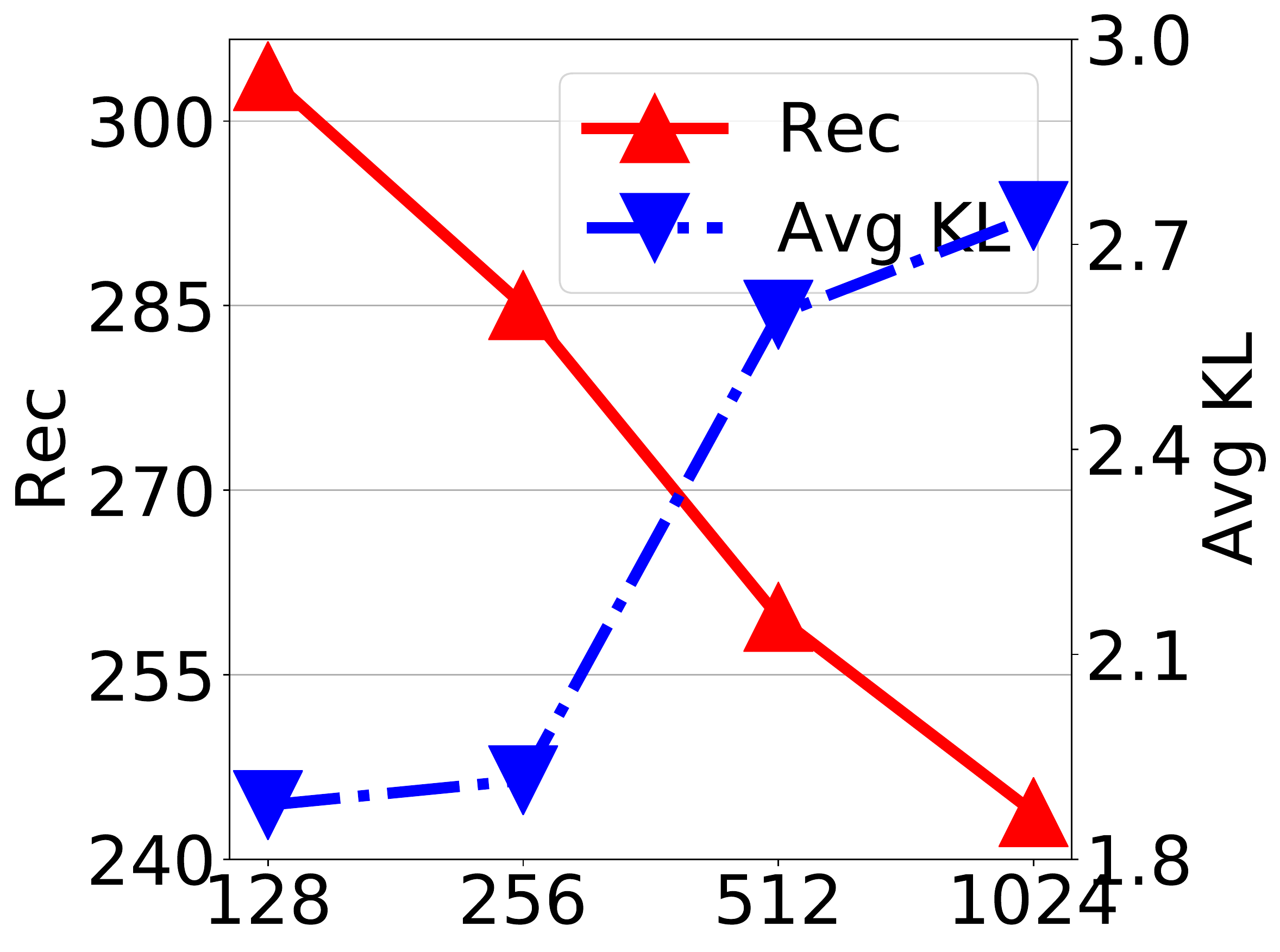}     
	}
	\subfigure[Max. regularizer $\beta_{\max}$]{
		\label{fig:vq_weight}
		\includegraphics[width=0.225\textwidth]{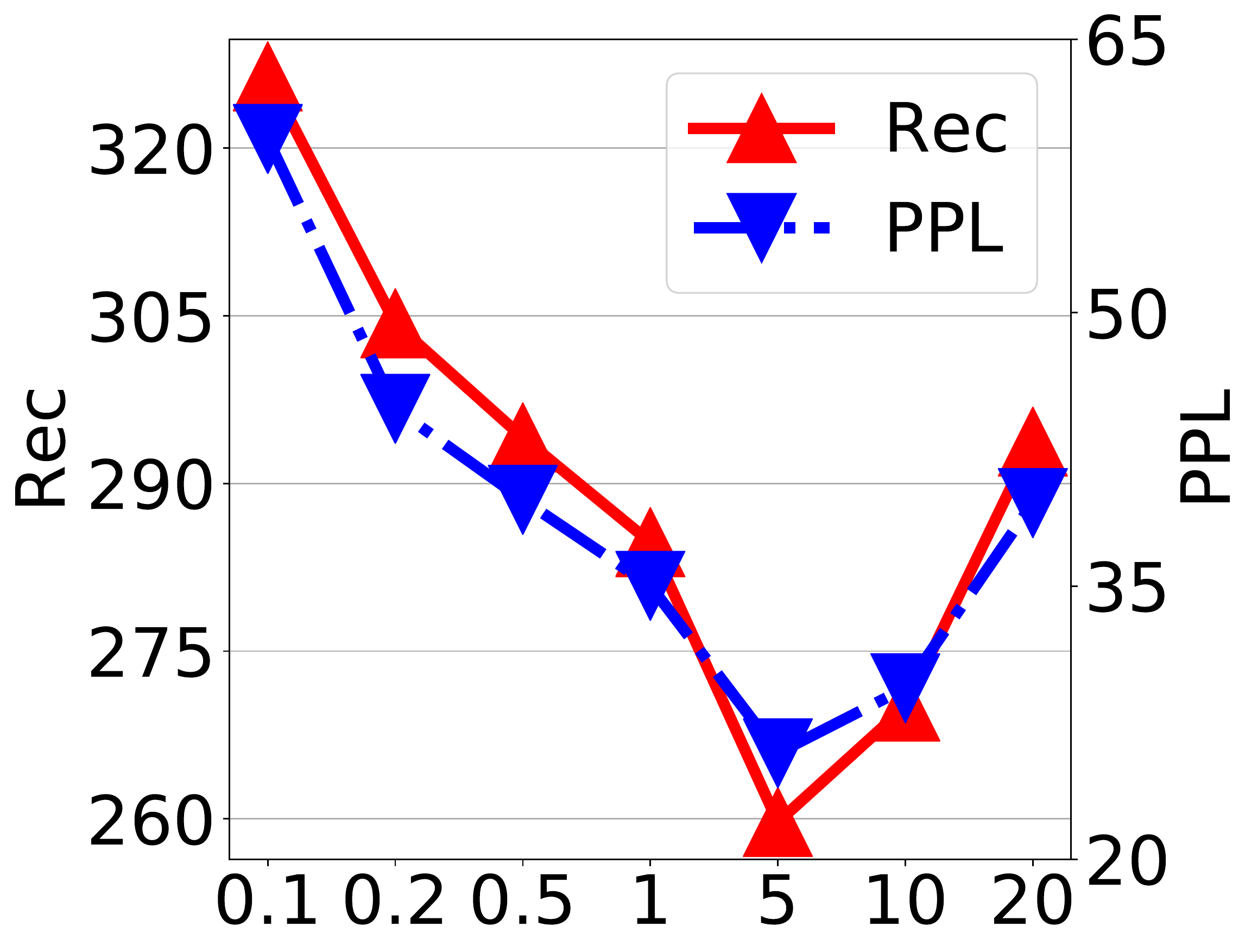}  
	}
	\subfigure[Latent dimension of $e_{z_t}$ ]{
		\label{fig:latent_dimension}
		\includegraphics[width=0.225\textwidth]{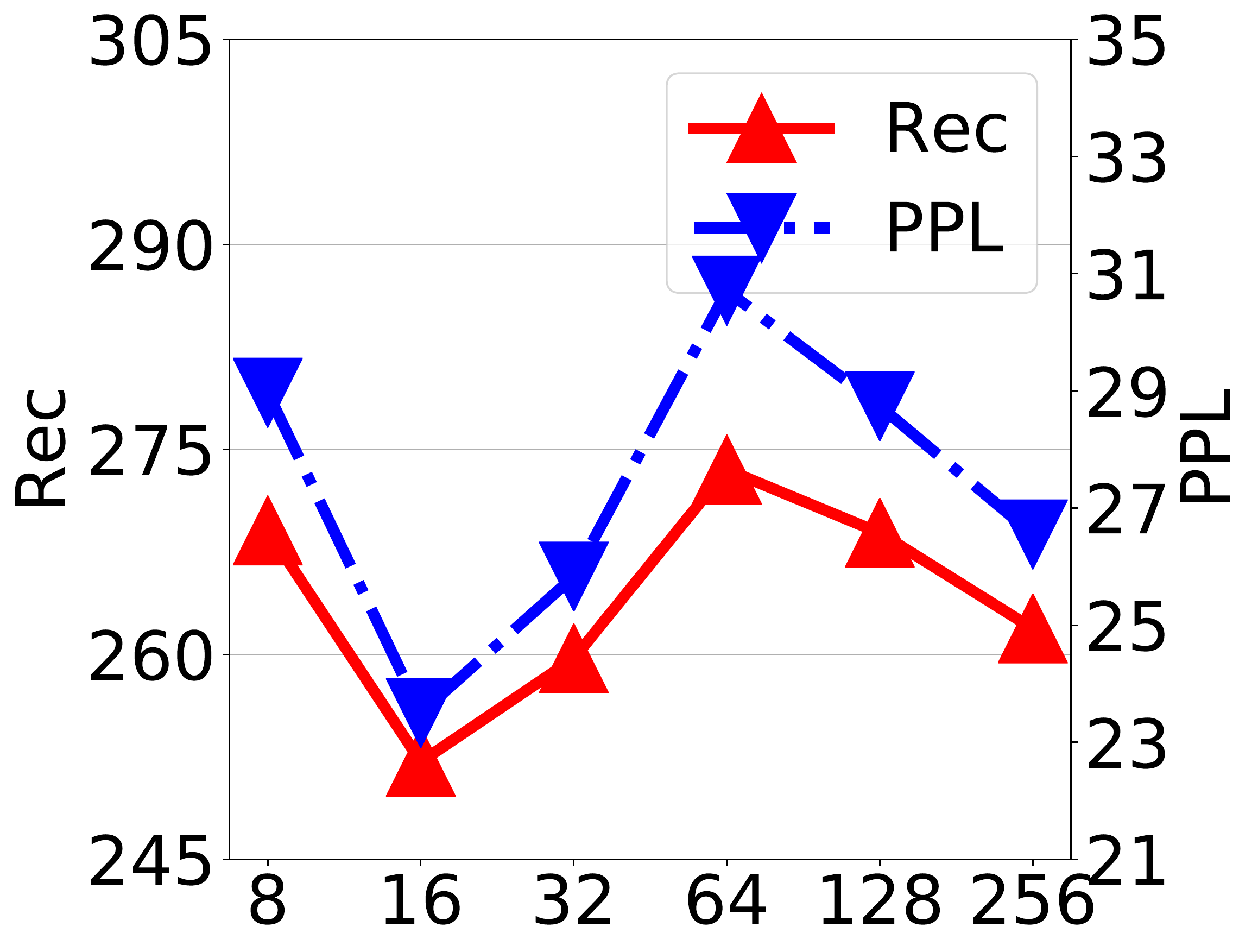}
	}
	\subfigure[Training dynamics]{
	    \label{fig:DAVAM_GAVAM_training}
	    \includegraphics[width=0.225\textwidth]{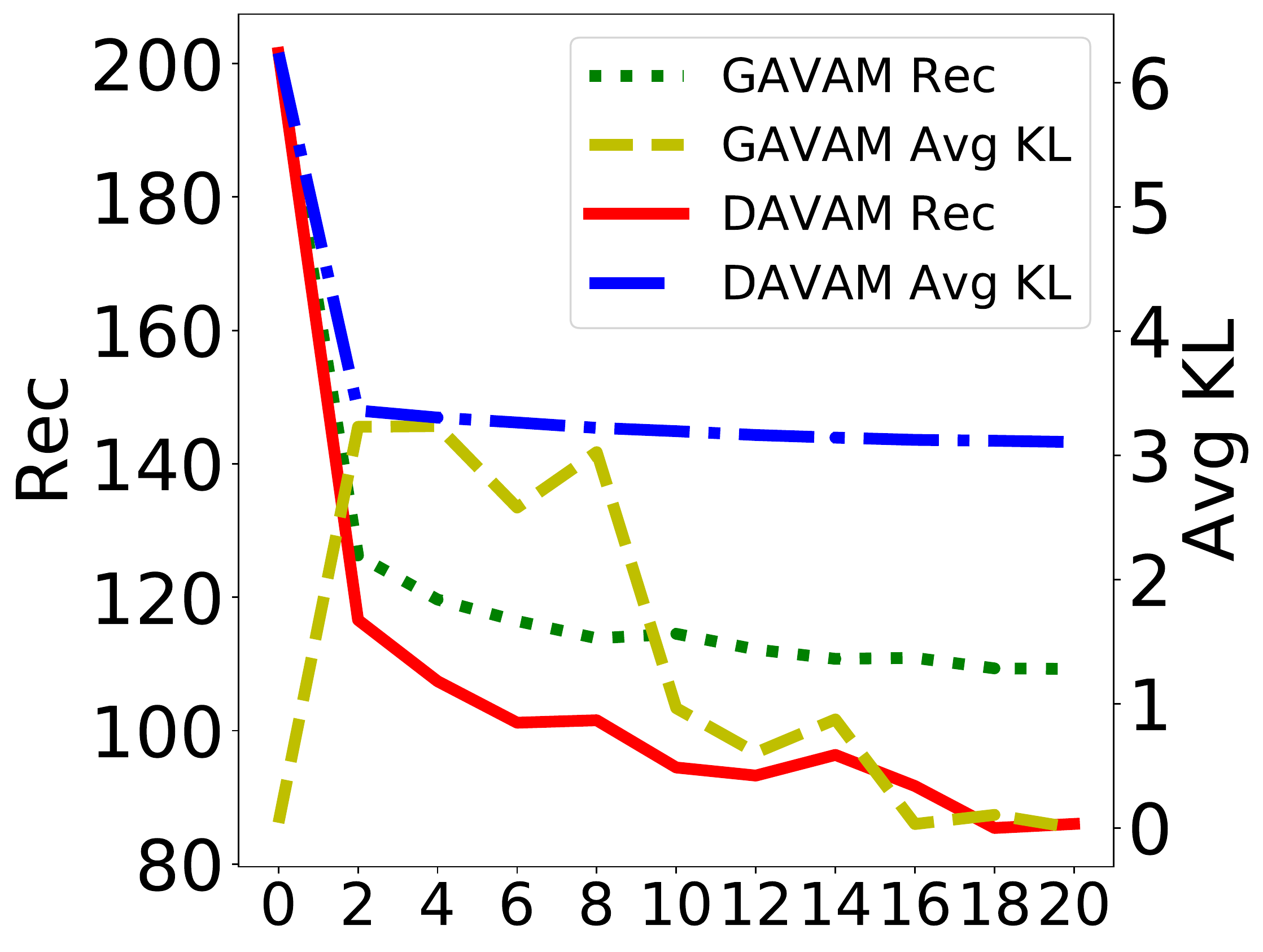}
	}
	\vspace{-0.5ex}
	\caption{Further analysis of DAVAM.}
	\vspace{-3ex}
\end{figure*}

To study the quality more rigorously,
we further sample 100 sentences randomly from VAE+Attn, GAVAM, and DAVAM with different lengths, and compare their average perplexity scores with the GPT-2 evaluator. 
As it is an inevitable trade-off between language modeling and generation, we also report the reconstruction loss on Yahoo.
The results are listed in Table~\ref{tab:gpt2_scores}.
It can be found that while VAE+Attn has a superior advantage in language modeling, it has the worst GPT-2 perplexity scores since i.i.d. Gaussian noises contain no sequential information.
GAVAM has minor improvement over VAE+Attn on GPT-2 scores thanks to the auto-regressive prior, but it performs poorly on language modeling due to posterior collapse. Finally, DAVAM generally achieves the lowest perplexity scores with reasonable ability in language modeling. This indicates the superiority of the auto-regressive prior for generation from scratch, and the power of discreteness to avoid posterior collapse in fitting observations.


\label{sec:exp_gq}

\subsubsection{Generation Diversity}
Diversity is another dimension to measure the success of language generation.
We follow~\cite{li2015diversity} to compute the entropy and the percentage of distinct unigrams or bigrams, which are denoted as Ent., Dist-1, and Dist-2 respectively. From Table~\ref{table:generation_diversity}, it can be observed that pretraining+FBP VAE achieves the highest diversity scores.
However, VAE+Attn, achieves the poorest diversity scores due to repeated words, as shown in Table~\ref{table:sentences}.
GAVAM has only minor improvement over VAE+Attn, and repeated words frequently occur as well. Finally, our DAVAM can generate diverse sentences despite the scores being slightly lower than pretraining+FBP VAE. This is due to the training of auto-regressive prior yields over-confident choices of latent codebooks, which can better capture sequential dependency with higher generation quality, but at the sacrifice of less generation diversity.









\subsubsection{Generation from Scratch as Data Augmentation}

Given the superior generation quality and diversity of DAVAM, we now apply it for data augmentation to improve language models that are trained over limited corpus. 
By amortizing the training instances into model parameters, DAVAM is able to generate sentences directly from random noises (i.e. generation from scratch).
Specifically, we selectively choose the corpus size in \{1k, 2k, 4k\} on SNLI dataset, and use a pre-trained DAVAM model to augment $\{0.5\times, 1\times, 2\times, 4\times\}$ times of th corpus. Figure~\ref{fig:barchart} shows the corresponding improvement of perplexity scores. It can be found that the perplexity decreases proportionally to the training size. Therefore, our DAVAM can be applied to language models that are trained with limited corpus for further improvement.

\subsection{Further Analysis}
\label{sec:exp_sa}

To gain a better understanding of our proposed DAVAM, we first conduct a set of sensitivity analysis on hyper-parameter settings of the model.
By default, all sensitivity analysis are conducted on Yahoo dataset with default parameter settings, except for the parameter under discussion. Then we turn to analyze training dynamics of DAVAM, which explains why DAVAM avoids posterior collapse.
\subsubsection{Code Book Size $K$}
We begin with the effect of different code book size $K$ on the reconstruction loss and KL divergence for language modeling. We vary $K\in\{128,256,512,1024\}$, and the results are shown in Figure~\ref{fig:two_stage}. It can be observed that as $K$ increases, the Rec loss decreases, whereas KL increases, both monotonically. The results are also consistent to Table~\ref{table:density_results} by increasing $K$ from 128 to 512.
Such phenomenons are intuitive since a larger $K$ improves model capacity but poses more challenges for training the auto-regressive prior.
Consequently, one should properly choose the code book size, such that the prior can approximate the posterior well, and yet the posterior is representative enough for the semantic dependency.




\subsubsection{Maximum Regularizer $\beta_{max}$}
Then we tune the maximum regularizer $\beta_{max}$, which controls the distance of the continuous hidden state $h_{1:T}^e$ to the code book $\{e_k\}_{k=1}^K$. Recall that a small $\beta_{max}$ loosely restricts the continuous space $h_{1:T}^e$ to the code book, making the quantization hard to converge. On the other hand, if $\beta_{max}$ is too large, $h_{1:T}^e$ could easily get stuck in some local minimal during the training. Therefore, it is necessary to find a proper trade-off between the two situations. We vary $\beta_{max} \in \{0.1, 0.2, 0.5, 1,5,10, 20\}$, and the results is shown in Figure~\ref{fig:vq_weight}.
We can find that when $\beta_{max}=5$, DAVAM achieves the lowest Rec, while smaller or larger $\beta_{max}$ both lead to higher Rec values.

\subsubsection{Dimension of Code Book Vectors}
Finally, we change dimension of $\{e_k\}_{k=1}^K$ in $\{8,16,32,64,128,256\}$, and the results are shown in Figure~\ref{fig:latent_dimension}. The performance of language modeling is relatively robust to the choice of the latent dimension. This is different from the continuous space where the dimension of latent variables is closely related to the model capacity. In the discrete scenario, the model capacity is largely determined by the code book size $K$ instead of the dimension of code book, which is also verified in Table~\ref{table:density_results} and Figure~\ref{fig:two_stage}.



\subsubsection{Training Dynamics}
\label{sec:exp_td}
To empirically understand how DAVAM addresses posterior collpase, we  investigate their training dynamics. We plot the curve of Rec and KL on the validation set of PTB in Figure~\ref{fig:DAVAM_GAVAM_training}.
We find the KL of GAVAM rises at the beginning to explain observations but diminishes quickly afterward. In the meanwhile, Rec does not decrease sufficiently. This shows that the collapsed posterior fails to explain the observations. 
For DAVAM, on the other hand, since the optimization of reconstruction is not affected by the KL divergence, Rec is minimized sufficiently in the first place. Then we minimize KL, which converges quickly without oscillation. In other words, the posterior and prior are updated separately in two stages to avoid posterior collapse.

%% file: sections/related.tex
\section{Related Work}
\subsection{Variational Attention Models}
Attention mechanism is commonly adopted address the issue of under-fitting in various deep generative models~\cite{kim2019attentive,deng2018latent,bahuleyan2017variational}.
Both \cite{deng2018latent} and \cite{bahuleyan2017variational} consider the generation from some source input, where new latent sequences are generated conditioned on observations. Nevertheless, these methods can hardly be applied when no source information is available. Instead, our work focuses on the ability of \textit{generation from scratch}, i.e., generating from latent space directly without external sources~\cite{subramani2019can}. Generation from scratch has various applications, such as data augmentation where new training instances can be directly generated from random noise to increase the limited training size. To enable such ability, an auto-regressive prior should be deployed to generate semantically dependent latent sequences. This explains the core idea of auto-regressive variational attention in our approach.

\subsection{Discrete Latent Variables}
Aside from the mostly used Gaussian distribution in VAEs, recent works also explore discrete latent space such as DVAE~\cite{Rolfe2017DiscreteVA}, DVAE++~\cite{Vahdat2018DVAEDV}, and DVAE\#~\cite{Vahdat2018DVAEDV1}.
Nevertheless, these works have different motivations for discreteness. They introduce binary latent variables to improve the model capacity. In DAVAM, instead of enhancing the model capacity, we assign one-hot distribution on latent variables that aims to resolve posterior collapse, which is not addressed in these previous efforts~\cite{Rolfe2017DiscreteVA,Vahdat2018DVAEDV,Vahdat2018DVAEDV1}.


%% file: sections/conclusion.tex
\section{Conclusion}
In this paper, we propose the discrete auto-regressive variational attention model, a new deep generative model for text modeling. The proposed approach addresses two important issues: information underrepresentation and posterior collapse. Empirical results on benchmark datasets demonstrate the superiority of our approach in both language modeling and auto-regressive generation.
While the proposed method focuses on the fundamental text modeling, it is also promising to extend to more applications such as machine translation~\cite{li2018multi,yang2019context,li2019information,li2020neuron,li2021diversity}, question answering~\cite{xu2020schema2qa}, log generation~\cite{he2017towards,he2016evaluation}, and code generation~\cite{li2017software,li2018code}.


%% file: sections/ack.tex
\section*{Acknowledgement}
The work described in this paper was partially supported by the Research Grants Council of the Hong Kong Special Administrative Region, China (CUHK 2410021, Research Impact Fund (RIF), R5034-18).